\documentclass[11pt,a4paper]{article}
\usepackage[hyperref]{emnlp-ijcnlp-2019}

\usepackage{times}
\usepackage{latexsym}

\usepackage{amsfonts}
\usepackage{url}
\usepackage{xcolor}
\usepackage[ruled,linesnumbered]{algorithm2e}
\usepackage{graphicx}
\usepackage{bbm}

\usepackage{amsmath}

\aclfinalcopy 

\title{Review Conversational Reading Comprehension}

\author{Hu Xu\textsuperscript{\text{1}}, Bing Liu\textsuperscript{\text{1}}, Lei Shu\textsuperscript{\text{1}}\and Philip S. Yu\textsuperscript{\text{1,2}}\\
    \textsuperscript{1}{Department of Computer Science, University of Illinois at Chicago, Chicago, IL, USA}\\
    \textsuperscript{2}{Institute for Data Science, Tsinghua University, Beijing, China}\\
    {\tt \{hxu48, liub, lshu3, psyu\}@uic.edu}
}

\date{}

\begin{document}
\maketitle
\begin{abstract}
Inspired by conversational reading comprehension (CRC), this paper studies a novel task of leveraging reviews as a source to build an agent that can answer multi-turn questions from potential consumers of online businesses. We first build a review CRC dataset and then propose a novel task-aware pre-tuning step running between language model (e.g., BERT) pre-training and domain-specific fine-tuning.~The proposed pre-tuning requires no data annotation, but can greatly enhance the performance on our end task. Experimental results show that the proposed approach is highly effective and has competitive performance as the supervised approach.\footnote{The dataset is available at \url{https://github.com/howardhsu/RCRC}.}


\end{abstract}
    
\section{Introduction}
\label{sec:intro}
Seeking information to assess whether a product or service suits one's needs is an important activity in consumer decision making.
One major hindrance for online businesses is that the consumers often have difficulty to get answers for their questions. 
With the ever-changing environment, it is very hard, if not impossible, for businesses to pre-compile an up-to-date knowledge base to answer user questions as in KB-QA \cite{kwok2001scaling,fader2014open,yin2015neural,xu2016question}.
Although community question-answering (CQA) helps~\cite{mcauley2016addressing}, one has to be lucky to get an existing customer to answer a question quickly. There is work on retrieving whole reviews relevant to a question~\cite{mcauley2016addressing,yu2018aware}, but it is not ideal for the user to read the whole reviews to fish for answers. 

\begin{table}
    \caption{An example of RCRC (best viewed in colors): a dialogue with 5 turns of customers' questions and answer spans from a review.}
    \centering
    \scalebox{0.75}{
        \begin{tabular}{l}
            \hline
            {\bf A Laptop Review:}\\
            \hline
            I purchased my Macbook Pro Retina from my school since I \\
            had a student discount , but I would gladly purchase it from \\
            Amazon for full price again if I had too . The Retina is \textcolor{purple}{\textbf{great}} \\
            , its \textcolor{red}{\textbf{amazingly fast}} when it boots up because of the \textcolor{orange}{\textbf{SSD}}\\
            \textcolor{orange}{\textbf{storage}} and the clarity of the screen is \textcolor{blue}{\textbf{amazing}} as well...\\
            \hline
            {\bf Turns of Questions from a Customer:}\\
            \hline
            \textcolor{purple}{$q_1$: how is retina display ?}\\
            \textcolor{red}{$q_2$: speed of booting up ?}\\
            \textcolor{orange}{$q_3$: why ?}\\
            \textcolor{cyan}{$q_4$: what 's the capacity of that ? (NO ANSWER)}\\
            \textcolor{blue}{$q_5$: is the screen clear ?}\\
            \hline
        \end{tabular}
    }
    \label{tbl:example}
    \vspace{-5mm}
\end{table}

Inspired by conversational reading comprehension (CRC) \cite{reddy2018coqa,choi2018quac,xu-etal-2019-bert}, we explore the possibility of turning reviews into a valuable source of knowledge of real-world experiences and using it to answer customer or user multi-turn questions.~We call this \textit{\underline{R}eview \underline{C}onversational \underline{R}eading \underline{C}omprehension} (RCRC).
The conversational setting enables the user to go into details via more specific questions and to simplify their questions by either omitting or co-referencing information in the previous context.
As shown in Table \ref{tbl:example}, 
the user first has an \textit{opinion} question about ``retina display'' (an \textit{aspect}) of a laptop.~Then he/she carries (or omits) the question type \textit{opinion} from the first question to the second question about another \textit{aspect} ``boot-up speed''.
Later, he/she carries the \textit{aspect} of the second question, but changes the question type to \textit{opinion reason} and then co-references the \textit{aspect} ``SSD'' from the third answer and asks for the capacity (a \textit{sub-aspect}) of ``SSD''.
Unfortunately, there is no answer in this review. 
Finally, the customer asks another \textit{aspect} as in the fifth question. RCRC is defined as follows.

\noindent\textbf{RCRC Definition}: Given a review that consists of a sequence of $n$ tokens $d=(d_1, \dots, d_n)$, a history of past $k-1$ questions and answers as the context $C=(q_1, a_1, q_2, a_2, \dots, q_{k-1}, a_{k-1})$ and the current question $q_k$, find a sequence of tokens (a textual span) $a=(d_s, \dots, d_e)$ in $d$ that answers $q_k$ based on $C$, where $1 \le s \le n$, $s\le e \le n$, and $s\le e$, or return \textit{NO ANSWER} ($s, e=0$) if the review does not contain the answer for $q_k$.

Note that although RCRC focuses on one review, it can potentially be deployed on the setting of multiple reviews (e.g., all reviews for a product), where the context $C$ may contain answers from different reviews.
To the best of our knowledge, there are no existing review datasets for RCRC. We first build a dataset called $(\text{RC})_2$ based on laptop and restaurant reviews from SemEval 2016 Task 5.\footnote{\url{http://alt.qcri.org/semeval2016/task5/} We choose this dataset to better align with existing research in sentiment analysis.} 
Given the wide spectrum of domains in online businesses and the prohibitive cost of annotation, $(\text{RC})_2$ has limited training data,
as in many other tasks of sentiment analysis. 

As a result, the challenge is how to effectively improve the performance of RCRC.
We adopt BERT \cite{devlin2018bert} as our base model since it can be either a feature encoder or a standalone model that achieves good performance on CRC \cite{reddy2018coqa}.
BERT bears with task-agnostic features, which require task-specific architecture and many supervised training examples to train(fine-tune) on an end task.
As $(\text{RC})_2$ has limited training data, 
we propose a novel task-aware \textit{pre-tuning} to further bridge the gap between BERT pre-training and RCRC task-awareness. 
Pre-tuning requires no annotation of CRC (or RCRC) data but just QA pairs (from CQA) and reviews that are largely available online.
The data are general and can potentially be used in other machine reading comprehension tasks.
Experimental results show that the proposed approach achieves competitive performance even compared with the supervised approach using a large-scale annotated dataset.

\section{BERT and Proposed Pre-tuning}
\label{sec:pt}

\subsection{BERT}
BERT is one of the key innovations \cite{peters2018deep,howard2018universal,radford2018improving,devlin2018bert} of unsupervised learning with the idea to learn contextualized representations that are previously only learned from supervised data.
It has two novel pre-training objectives that greatly improve the learned representations: masked language model (MLM) and next sentence prediction (NSP). The former predicts a randomly masked word and the latter tries to detect whether two sides of a text are from the same document or not. A training example is formulated as $(\texttt{[CLS]}, x_{1:j}, \texttt{[SEP]}, x_{j+1:n}, \texttt{[SEP]})$, where \texttt{[CLS]} is a special token and $x_{1:n}$ is a document (with randomly masked words) split into two sides $x_{1:j}$ and $x_{j+1:n}$ and \texttt{[SEP]} separates those two.
We leverage BERT's architecture and focus on how to design a textual format to bring task-awareness of RCRC into a (pre-tuned) model.

\subsection{Textual Format}
\label{sec:format}
Inspired by the DrQA system \cite{reddy2018coqa}, we formulate an input example $x$ for both RCRC fine-tuning and pre-tuning\footnote{We share the same notation for both tasks for brevity.} as a composition of the context $C$, the current question $q_k$, and a review $d$:
\vspace*{-2mm}
\begin{equation}
\begin{split}
(\texttt{[CLS]} \texttt{[Q]} q_1 \texttt{[A]} a_1 \dots \texttt{[Q]} q_{k-1} \texttt{[A]} a_{k-1} \\
\texttt{[Q]} q_{k} \texttt{[SEP]} d_{1:n} \texttt{[SEP]}),
\nonumber
\end{split}
\vspace*{-5mm}
\end{equation}
where past QA pairs $q_1, a_1, \dots, q_{k-1}, a_{k-1}$ in $C$ are concatenated and separated by two tokens \texttt{[Q]} and \texttt{[A]} and then concatenated with the current question $q_k$ as the left side of BERT and the right side is the review document. 
One can observe that BERT lacks the basic understanding of the RCRC task regarding both the input and output, such as the above input format 
and textual spans in a review. Limited training data of $(\text{RC})_2$ may not be sufficient to learn such a complex input and output.
We propose a pre-tuning stage that can enhance the understanding of the input/output before fine-tuning on $(\text{RC})_2$.

\subsection{Data Formulation for Pre-tuning}
\label{sec:form}
We first formulate the data for pre-tuning that aims to address the understanding of the textual format.
As we have no annotated data except the limited $(\text{RC})_2$ data, we harvest domain QA pairs and reviews (that are largely available online, see Section \ref{sec:exp}), which are typically organized under an entity (a laptop or a restaurant). The QA pairs and reviews are combined to produce the pre-tuning examples. The process is given in Algorithm \ref{alg:pre-tuning}.

\begin{algorithm}[t]
    \LinesNumbered
    \DontPrintSemicolon
    \caption{Data Generation Algorithm}
    \label{alg:pre-tuning}
    \SetKwInOut{Input}{Input} 
    \SetKwInOut{Output}{Output} 
    \Input{$\mathcal{Q}$: a set of QA pairs;\\$\mathcal{R}$: a set of reviews;\\$h_{max}$: maximum turns in context.}
    \Output{$\mathcal{T}$: pre-tuning data.}

    \BlankLine

    $\mathcal{T} \gets \{\}$ \;
    \For{$(q', a') \in \mathcal{Q}$ }{
        $x \gets \texttt{[CLS]} $ \;
        $h \gets \text{RandInteger}([0, h_{\text{max}}]) $ \;
        \For{$1 \to h$}{
            $q'', a'' \gets \text{RandSelect}(\mathcal{Q}\backslash(q', a'))$ \;
            $ x \gets x \oplus \texttt{[Q]} \oplus q'' \oplus \texttt{[A]} \oplus a'' $\;
        }
        $ x \gets x \oplus \texttt{[Q]} \oplus q' \texttt{[SEP]} $ \;
        $ r_{1:m} \gets \text{RandSelect}(\mathcal{R}) $ \;
        \If{$\text{RandFloat}([0.0, 1.0]) > 0.5$ }{
            $(\_, a) \gets \text{RandSelect}(\mathcal{Q}\backslash(q', a') ) $ \;
            $(u, v) \gets (1, 1)$ \;
        }
        \Else{$a \gets a'$ \;
            $(u, v) \gets (|x|, |x|+|a|) $ \;
        }
        $ l \gets \text{RandInteger}([0, u]) $ \;
        $ d_{1:n} \gets r_{0:l} \oplus a \oplus r_{l+1:u} $ \;
        \If{$u>1$ }{
            $ (u, v) \gets (u+|r_{0:l}|, v+|r_{0:l}|) $
        }
        $x \gets x \oplus d_{1:n} \oplus \texttt{[SEP]}$ \;
        $\mathcal{T} \gets \mathcal{T} + (x, (u, v) )$ \;
    }
\end{algorithm}

The inputs to Algorithm \ref{alg:pre-tuning} are a set of QA pairs and a set of reviews belonging to the same entity and the maximum number of turns in the context. The output is the pre-tuning data, which is initialized in Line 1. 
Each example is denoted as $(x, (u, v))$, where $x$ is the input example and $(u, v)$ indicates the boundary (starting and ending indexes) of an answer for the auxiliary objective (discussed in Section \ref{sec:aux}).
Given a QA pair $(q', a')$ in Line 2, we first build the left side of input example $x$ in Line 3-9.
After initializing input $x$ in Line 3, we randomly determine the number of turns in the context in Line 4 and concatenate $\oplus$ these turns of QA pairs in Line 5-8, where $\mathcal{Q}\backslash(q', a')$ ensures the current QA pair $(q', a')$ is not chosen.
In Line 9, we concatenate with the current question $q'$.
Lines 10-23 build the right side of input example $x$ and the answer boundary.
In Line 10, we randomly draw a review $r$ with $m$ sentences. To challenge the pre-tuning stage to discover the semantic relatedness between $q'$ and $a'$ (for the auxiliary objective), we first decide whether to allow the right side of $x$ contains $a'$ (Line 16) for $q'$ or a random (negative/no) answer $a$ in Lines 11-12.
We also come up with two indexes $u$ and $v$ initialized in Lines 13 and 17.
Then, we insert $a$ into review $r$ by randomly picking one from the $m+1$ locations in Lines 19-20.
This gives us $d_{1:n}$, which has $n$ tokens. 
We further update $u$ and $v$ to allow them to point to the chunk boundaries of $a'$.
Otherwise, BERT should detect no $a'$ on the right side and point to \texttt{[CLS]} ($u,v=1$). Finally, examples are aggregated in Line 25.
Algorithm \ref{alg:pre-tuning} is run $k$ times to allow for enough samplings. 
Following BERT, we still randomly mask some words in each example $x$ but omitted here for brevity.

\subsection{Auxilary Objective}
\label{sec:aux}
Besides the input, we further adapt BERT to the output of RCRC with an auxiliary objective.
The design of this auxiliary objective is to mimic a prediction of a textual span in RCRC, which aims to predict the token spans of an answer randomly inserted in the review or \textit{NO ANSWER} if a randomly drawn negative answer appears.
The implementation of both the auxiliary objective and RCRC model is similar to BERT for SQuAD 2.0 \cite{rajpurkar2018know}, so we omit them for brevity.
After pre-tuning, we fine-tune using the $(\text{RC})_2$ dataset to show the performance of RCRC. 

\section{Experiments}
\label{sec:exp}

We adopt SemEval 2016 Task 5 
as the review source for RCRC (to be consistent with research in sentiment analysis), which contains two domains \textit{laptop} and \textit{restaurant}.
We kept the split of training and testing, and annotated dialogues on each review.
Similar to the way of annotating the CoQA data \cite{reddy2018coqa}, we first let both annotators read a review and then one annotator asks questions and the other annotator annotates the answer token spans (or no answer)\footnote{The annotated data is in the format of CoQA \cite{reddy2018coqa} to help future research. But we do not focus on generative annotation as in CoQA because businesses are sensitive to errors of generative models}.
To ensure questions are real-world questions, annotators are first asked to read hundreds of community questions and answers (CQA) from real customers.
Since the number of testing reviews is small, we encourage annotators to produce as many dialogues as possible. 
Each training review is encouraged to have 1 or 2 dialogues.
The statistics of the annotated $(\text{RC})_2$ dataset is shown in Table \ref{tbl:rcrc}.
We use 20\% of the training reviews as the validation set for each domain.

\begin{table}
    \caption{Statistics of $(\text{RC})_2$ Datasets.}
    \centering
    \scalebox{0.8}{
        \begin{tabular}{c||c|c}
        \hline
        Training &{\bf Laptop } &{\bf Restaurant} \\
        \hline
        \# of reviews & 445 & 350\\
        \# of dialogues & 506 & 382\\
        \# of dialog /w 3+ turns & 375 & 315\\
        \# of questions & 1679 & 1486\\
        \% of no answers & 24.3\%& 24.2\%\\
        \hline
        Testing &{\bf Laptop } &{\bf Restaurant} \\
        \hline
        \# of reviews & 79 & 90 \\
        \# of dialog & 170 & 160 \\
        \# of dialog /w 3+ turns & 148 & 135\\
        \# of questions & 804 & 803 \\
        \% of no answers & 26.6\% & 28.0\% \\
        \hline
        \end{tabular}
    }
\label{tbl:rcrc}
\vspace{-5mm}
\end{table}

For the proposed pre-tuning, we collect QA pairs and reviews for these two domains.
For \emph{laptop}, we collect the reviews from \cite{he2016ups} and QA pairs from \cite{Xu2018pro} both under the laptop category of Amazon.com. We exclude products in the test data of $(\text{RC})_2$. 
This gives us 113,728 laptop reviews and 19,104 QA pairs. 
For \emph{restaurant}, we crawl reviews and all QA pairs from the top 60 restaurants in each U.S. city from Yelp.com.
This ends with 197,333 restaurant reviews and 49,587 QA pairs. Based on the number of QAs, Algorithm 1 is run $k=10$ times for laptop and $k=5$ times for restaurant.

To compare with the performance of a fully-supervised approach,~we leverage the CoQA dataset with 7,199 documents (covering domains in Children’s Story, Mid/High School Literature, News, Wikipedia, etc.) and 108,647 turns of question/answer span annotated via crowdsourcing.

We compare the following methods: 
\textbf{DrQA} is a CRC baseline coming with the CoQA dataset\footnote{https://github.com/stanfordnlp/coqa-baselines}. 
\textbf{DrQA+CoQA} is the above baseline pre-tuned on the CoQA dataset and then fine-tuned on $(\text{RC})_2$ to show that even DrQA pre-trained on CoQA is sub-optimal.
\textbf{BERT}\footnote{We choose $\text{BERT}_{\text{BASE}}$ as we cannot fit $\text{BERT}_{\text{LARGE}}$ into the memory.} is the pre-trained BERT weights directly fine-tuned on $(\text{RC})_2$ for ablation study on the effectiveness of pre-tuning.
\textbf{BERT+review} first tunes BERT on domain reviews using the same objectives as BERT pre-training and then fine-tunes on $(\text{RC})_2$. We use this baseline to show that a simple domain-adaptation of BERT is not sufficient.
\textbf{BERT+CoQA} first fine-tunes BERT on the supervised CoQA data and then fine-tunes on $(\text{RC})_2$. We use this baseline to show that even compared with using this large-scale supervised data, our pre-tuning is still very competitive.
\textbf{BERT+Pre-tuning} is the proposed approach.

We set the maximum length of BERT to 256 with the maximum length of context+question to 96 ($h_\text{max}=9$ for Algorithm \ref{alg:pre-tuning}) and the batch size to 16.
We perform pre-tuning for 10k steps. 
CoQA fine-tuning converges in 2 epochs.
Fine-tune RCRC is performed for 4 epochs and most runs converged within \textbf{3} epochs. 
We search the maximum number of turns in context $C$ for RCRC fine-tuning using the validation set, which ends with 6 turns for laptop and 5 turns for restaurant.
Results are reported as averages of 3 runs. 
To be consistent, we leverage the same evaluation script as CoQA, 
which reports 
turn-level Exact Match (EM) and F1 scores for all turns in all dialogues.


\begin{table}
    \caption{RCRC on EM (Exact Match) and F1.}
    \centering
    \scalebox{0.7}{
        \begin{tabular}{l||c c|c c}
        \hline
        {\bf Domain} & {\bf Laptop} & & {\bf Rest.} & \\
        \hline
        {\bf Methods} & {\bf EM } &{\bf F1 } & {\bf EM } & {\bf F1 } \\
        \hline
        DrQA & 28.5 & 36.6 & 41.6 & 50.3 \\
        DrQA+CoQA(supervised) & 40.4 & 51.4 & 47.7 & 58.5 \\
        \hline
        \hline
        BERT & 38.57 & 48.67 & 46.87 & 55.07 \\
        BERT+review & 34.53 & 43.83 & 47.23 & 53.7 \\
        BERT+CoQA(supervised) & 47.1 & 58.9 & 56.57 & 67.97 \\
        BERT+Pre-tuning & 46.0 & 57.23 & 54.57 & 64.43 \\
        \hline
        \end{tabular}
    }
\label{tbl:result_rcrc}
\vspace{-7mm}
\end{table}

\noindent\textbf{Result Analysis}\\
As shown in Table \ref{tbl:result_rcrc}, 
BERT+Pre-tuning has significant performance gains over BERT fine-tuned directly on $(\text{RC})_2$ by \textbf{9\%}.
BERT is overall better than DrQA.
But directly using review documents to adapt BERT does not yield better results as in BERT+review.
We suspect the task of RCRC still requires a certain degree of general language understanding on the question side and BERT+review also has the effect of (catastrophic) forgetting \cite{kirkpatrick2017overcoming} on such representation.
Further, large-scale annotated CoQA data can boost the performance for both DrQA and BERT.
However, our pre-tuning approach still has competitive performance and it requires no annotation at all.
We examine the errors of BERT+Pre-tuning and realize that both locations of span and span boundaries tend to have errors, indicating a significant room for improvement.

\section{Conclusions}
In this paper, we aimed to build a novel review-based conversational reading agent using limited annotated data.
We proposed a (language-model like) pre-tuning method without requiring any other annotation and empirical results showed its effectiveness.

\bibliography{emnlp-ijcnlp-2019}

\begin{thebibliography}{17}
\expandafter\ifx\csname natexlab\endcsname\relax\def\natexlab#1{#1}\fi

\bibitem[{Choi et~al.(2018)Choi, He, Iyyer, Yatskar, Yih, Choi, Liang, and
  Zettlemoyer}]{choi2018quac}
Eunsol Choi, He~He, Mohit Iyyer, Mark Yatskar, Wen-tau Yih, Yejin Choi, Percy
  Liang, and Luke Zettlemoyer. 2018.
\newblock Quac: Question answering in context.
\newblock \emph{arXiv preprint arXiv:1808.07036}.

\bibitem[{Devlin et~al.(2018)Devlin, Chang, Lee, and
  Toutanova}]{devlin2018bert}
Jacob Devlin, Ming-Wei Chang, Kenton Lee, and Kristina Toutanova. 2018.
\newblock Bert: Pre-training of deep bidirectional transformers for language
  understanding.
\newblock \emph{arXiv preprint arXiv:1810.04805}.

\bibitem[{Fader et~al.(2014)Fader, Zettlemoyer, and Etzioni}]{fader2014open}
Anthony Fader, Luke Zettlemoyer, and Oren Etzioni. 2014.
\newblock Open question answering over curated and extracted knowledge bases.
\newblock In \emph{Proceedings of the 20th ACM SIGKDD international conference
  on Knowledge discovery and data mining}, pages 1156--1165. ACM.

\bibitem[{He and McAuley(2016)}]{he2016ups}
Ruining He and Julian McAuley. 2016.
\newblock Ups and downs: Modeling the visual evolution of fashion trends with
  one-class collaborative filtering.
\newblock In \emph{proceedings of the 25th international conference on world
  wide web}, pages 507--517. International World Wide Web Conferences Steering
  Committee.

\bibitem[{Howard and Ruder(2018)}]{howard2018universal}
Jeremy Howard and Sebastian Ruder. 2018.
\newblock Universal language model fine-tuning for text classification.
\newblock In \emph{Proceedings of the 56th Annual Meeting of the Association
  for Computational Linguistics (Volume 1: Long Papers)}, volume~1, pages
  328--339.

\bibitem[{Kirkpatrick et~al.(2017)Kirkpatrick, Pascanu, Rabinowitz, Veness,
  Desjardins, Rusu, Milan, Quan, Ramalho, Grabska-Barwinska
  et~al.}]{kirkpatrick2017overcoming}
James Kirkpatrick, Razvan Pascanu, Neil Rabinowitz, Joel Veness, Guillaume
  Desjardins, Andrei~A Rusu, Kieran Milan, John Quan, Tiago Ramalho, Agnieszka
  Grabska-Barwinska, et~al. 2017.
\newblock Overcoming catastrophic forgetting in neural networks.
\newblock \emph{Proceedings of the national academy of sciences}, page
  201611835.

\bibitem[{Kwok et~al.(2001)Kwok, Etzioni, and Weld}]{kwok2001scaling}
Cody Kwok, Oren Etzioni, and Daniel~S Weld. 2001.
\newblock Scaling question answering to the web.
\newblock \emph{ACM Transactions on Information Systems (TOIS)},
  19(3):242--262.

\bibitem[{McAuley and Yang(2016)}]{mcauley2016addressing}
Julian McAuley and Alex Yang. 2016.
\newblock Addressing complex and subjective product-related queries with
  customer reviews.
\newblock In \emph{Proceedings of the 25th International Conference on World
  Wide Web}, pages 625--635. International World Wide Web Conferences Steering
  Committee.

\bibitem[{Peters et~al.(2018)Peters, Neumann, Iyyer, Gardner, Clark, Lee, and
  Zettlemoyer}]{peters2018deep}
Matthew~E Peters, Mark Neumann, Mohit Iyyer, Matt Gardner, Christopher Clark,
  Kenton Lee, and Luke Zettlemoyer. 2018.
\newblock Deep contextualized word representations.
\newblock \emph{arXiv preprint arXiv:1802.05365}.

\bibitem[{Radford et~al.()Radford, Narasimhan, Salimans, and
  Sutskever}]{radford2018improving}
Alec Radford, Karthik Narasimhan, Tim Salimans, and Ilya Sutskever.
\newblock Improving language understanding by generative pre-training.

\bibitem[{Rajpurkar et~al.(2018)Rajpurkar, Jia, and Liang}]{rajpurkar2018know}
Pranav Rajpurkar, Robin Jia, and Percy Liang. 2018.
\newblock Know what you don't know: Unanswerable questions for squad.
\newblock \emph{arXiv preprint arXiv:1806.03822}.

\bibitem[{Reddy et~al.(2018)Reddy, Chen, and Manning}]{reddy2018coqa}
Siva Reddy, Danqi Chen, and Christopher~D Manning. 2018.
\newblock Coqa: A conversational question answering challenge.
\newblock \emph{arXiv preprint arXiv:1808.07042}.

\bibitem[{Xu et~al.(2019)Xu, Liu, Shu, and Yu}]{xu-etal-2019-bert}
Hu~Xu, Bing Liu, Lei Shu, and Philip Yu. 2019.
\newblock \href {https://doi.org/10.18653/v1/N19-1242} {{BERT} post-training
  for review reading comprehension and aspect-based sentiment analysis}.
\newblock In \emph{Proceedings of the 2019 Conference of the North {A}merican
  Chapter of the Association for Computational Linguistics: Human Language
  Technologies, Volume 1 (Long and Short Papers)}, pages 2324--2335,
  Minneapolis, Minnesota. Association for Computational Linguistics.

\bibitem[{Xu et~al.(2018)Xu, Xie, Shu, and Yu}]{Xu2018pro}
Hu~Xu, Sihong Xie, Lei Shu, and Philip~S. Yu. 2018.
\newblock Dual attention network for product compatibility and function
  satisfiability analysis.
\newblock In \emph{Proceedings of AAAI Conference on Artificial Intelligence
  (AAAI)}.

\bibitem[{Xu et~al.(2016)Xu, Reddy, Feng, Huang, and Zhao}]{xu2016question}
Kun Xu, Siva Reddy, Yansong Feng, Songfang Huang, and Dongyan Zhao. 2016.
\newblock Question answering on freebase via relation extraction and textual
  evidence.
\newblock \emph{arXiv preprint arXiv:1603.00957}.

\bibitem[{Yin et~al.(2015)Yin, Jiang, Lu, Shang, Li, and Li}]{yin2015neural}
Jun Yin, Xin Jiang, Zhengdong Lu, Lifeng Shang, Hang Li, and Xiaoming Li. 2015.
\newblock Neural generative question answering.
\newblock \emph{arXiv preprint arXiv:1512.01337}.

\bibitem[{Yu and Lam(2018)}]{yu2018aware}
Qian Yu and Wai Lam. 2018.
\newblock Aware answer prediction for product-related questions incorporating
  aspects.
\newblock In \emph{Proceedings of the Eleventh ACM International Conference on
  Web Search and Data Mining}, pages 691--699. ACM.

\end{thebibliography}
\bibliographystyle{acl_natbib}

\appendix




\end{document}